# Multi-Granularity Prompts for Topic Shift Detection in Dialogue

Jiangyi Lin, Yaxin Fan, Xiaomin Chu, Peifeng Li[✉] and Qiaoming Zhu

School of Computer Science and Technology, Soochow University, Suzhou, China
{jylin, yxfansuda}@stu.suda.edu.cn
{xmchu, pfli, qmzhu}@suda.edu.cn

**Abstract.** The goal of dialogue topic shift detection is to identify whether the current topic in a conversation has changed or needs to change. Previous work focused on detecting topic shifts using pre-trained models to encode the utterance, failing to delve into the various levels of topic granularity in the dialogue and understand dialogue contents. To address the above issues, we take a prompt-based approach to fully extract topic information from dialogues at multiple-granularity, i.e., label, turn, and topic. Experimental results on our annotated Chinese Natural Topic Dialogue dataset CNTD and the publicly available English TIAGE dataset show that the proposed model outperforms the baselines. Further experiments show that the information extracted at different levels of granularity effectively helps the model comprehend the conversation topics.

## 1 Introduction

The topic structure explains the topical relationship between two consecutive text units (e.g., paragraphs in a discourse, turns in a dialogue). As one of the essential dialogue analysis tasks, dialogue topic shift detection refers to detecting whether a topic shift has occurred in the response of a dialogue, which can help dialogue systems to change topics and actively guide the dialogue. Since this task can help variant models to understand dialogue topics, it is of great benefit for many downstream tasks, such as response generation [1] and reading comprehension [2,3]. It also can help those real-time applications to generate topics that perform well in dialogue scenarios due to its shift response [4,5,6].

The goal of dialogue topic shift detection is to identify the topic of dialogue by taking into account the current response information in real time. This task is similar to dialogue topic segmentation [7]. However, dialogue topic shift detection is more challenging than dialogue topic segmentation. In dialogue topic segmentation, all utterances are visible to each other, allowing the model to access the whole content of the dialogue after the responses have been given. However, dialogue topic shift detection is a real-time task and cannot access future utterances. For example in Fig. 1, there is a dialogue with two topics (e.g., favorite animal and weakness). The task of topic segmentation is to split this dialogue into two blocks, which can access all six utterances in this dialogue. The task of topic shift detection is to predict whether the next utterance will change the topic, based on all existing utterances. If we want to predict whether the topic is shifted during $u_1$ and $u_2$, we can only access two utterances, i.e., $u_1$ and $u_2$.

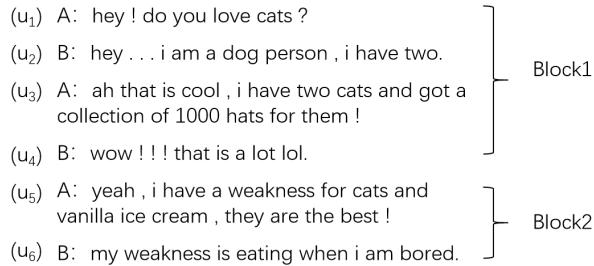

**Fig. 1.** An example of the topic structure in a dialogue of six utterances (i.e., $u_1$- $u_6$) where each block refers to a topic.

The majority of prior research on topic segmentation has focused on enhancing the model and disregarding the comprehensive exploration of information. As a result of insufficient training data, unsupervised techniques remain the prevailing choice for segmenting dialogue topics [8]. On the contrary, dialogue topic shift detection is a relatively new task in the field of dialogue topics. Although those topic segmentation models can be adapted in topic shift detection, the absence of future utterances makes it harder to distinguish the topic shift between utterances.

Only a few studies focused on dialogue topic shift detection [9-11]. Current studies on dialogue topics shift detection only focus on extracting surface semantic information using pre-trained models, without delving into deeper topic information. These models struggle to comprehend natural dialogues that involve randomness. To address this issue, we employ a prompt-based approach to extract dialogue information at multiple-granularity.

Moreover, classification and generative models can offer complementary benefits. While classification models tend to perform well in large categories due to their limited search space, generative models can incorporate prior knowledge to better understand small categories, leading to more natural language expressions with clearer explanations. Inspired by this trend, we combine classification and generation to enhance our final classification, while the generation model comprehends the conversation topics at three different levels of granularity. Specifically, we first match the original topic labels to the relevant target sentences. Then, we perform a thorough keyword extraction for each topic block and use these keywords to identify the topics within the block. Lastly, we apply semantic role labeling (SRL) on the dialogue sentences to create the target sentences. We annotated a Chinese Natural Topic Dialogue corpus CNTD in previous work[12] based on NaturalConv[13]. Experimental results on our annotated corpus CNTD dataset and the publicly available English TIAGE dataset show that the proposed model outperforms the baselines.

## 2 Background

We first briefly introduce the relevant dialogue topic corpus, then summarize the existing methods for dialogue topic detection, and finally introduce the related research on Prompt.

For English, Xie et al. [9] annotated the TIAGE corpus consisting of 500 dialogues with 7861 turns based on PersonaChat [14]. Xu et al. [11] built a dataset including 711 dialogues by joining dialogues from existing multi-turn dialogue datasets: MultiWOZ Corpus [15], and Stanford Dialog Dataset [16]. Both corpora are either small or limited to a particular domain, and neither applies to the study of the natural dialogue domain. For Chinese, Xu et al. [11] annotated a dataset including 505 phone records of customer service on banking consultation. However, this corpus is likewise restricted to a few specialized domains while natural dialogues are more complicated. Therefore, we annotated a Chinese Natural Topic Dialogue corpus CNTD in previous work[12], which contains 1308 natural conversations from six different domains. And we developed a benchmark on the response-unkown dialogue topic detection task. Current studies on dialogue topics shift detection only focus on extracting surface semantic information using pre-trained models, without delving into deeper topic information.

The field of detecting topic shifts in dialogue is still in its infancy and has received limited attention thus far. As we mentioned above, dialogue topic shift detection is similar to topic segmentation, we first discuss the related work in this area. Historically, due to the lack of training data, early studies in dialogue topic segmentation utilized unsupervised methods relying on word co-occurrence statistics [17] or sentence topic distributions [18] to determine sentence similarity between conversational turns and identify changes in thematic or semantic content. However, with the advent of large-scale corpora such as Wikipedia, supervised methods for monologic topic segmentation have gained popularity, especially those using neural-based approaches [19,20]. These supervised techniques have become the favored choice among researchers due to their improved performance and efficiency.

Dialogue topic shift detection is strongly different from dialogue topic segmentation. For the dialogue topic shift detection task, Xie et al. [9] are the first to define this task and predicted the topic shift based on the T5 model. In general, the dialogue topic shift detection task is still a challenge, as it can only rely on the context information of the dialogue. In this paper, based on a classification module, we use a generation module to further mine information from conversations, and joint training to facilitate real-time topic shift detection.

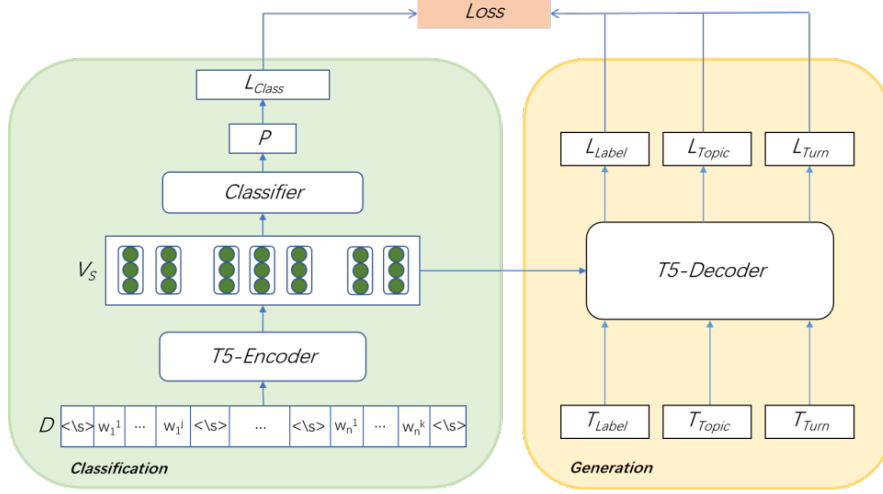

**Fig. 2.** Model architecture, which contains a classification module (left) and a generation module (right). We add the classification layer after the encoder to form the classification module. And the original decoder is used as the generation module to mine the conversation information by multiple-granularity, where $T$ denotes the target sentence and $P$ denotes the classification result.

## 3 Model

Our framework is shown in Fig. 2. To enhance the performance of topic detection in the classification module, we introduce a method that mines dialogue information from various granularities through the generation module. Our model is comprised of two sub-modules: the classification module and the generation module. Our classification module includes a classification layer at the end of the encoding layer. Moreover, we consider the task of dialogue topic shift detection as a text-generation task to achieve a more thorough understanding of the topics. Unlike previous studies, we employ a generative model from three granularities to produce the target sentences.

Finally, the generated results are mapped to the relevant relations to determine the relation types.

### 3.1 Classification Module

Let $C = \{du_1, \ldots, du_i, \ldots, du_{n-1}\}$ represents a set of existing utterances, where $n-1$ is the number of the existing utterances, and $du_i$ is the i-th utternace. Let $R = \{du_n\}$ represents a response utterance after $C$. Finally, the set of all known utterances includes $C$ and $R$, which can be denoted as $DU = \{C, R\}$.

We need to learn a model $f: DU \rightarrow Y(R)$ to classify the response utterance $R$ (i.e., $du_n$) into the predefined categories $Y = \{0,1\}$, which is the ground-truth label (0 denotes non-shift and 1 denotes shift).

Similar to the input of traditional classification models, we first convert $DU$ into a string $D$ as follows.

$$D = <\backslash s> du_1 <\backslash s> \ldots <\backslash s> du_n <\backslash s> \quad (1)$$

where $du_1 = \{w_1^1, \ldots, w_1^j\}$ and $du_n = \{w_n^1, \ldots, w_n^k\}$ denote the sequence of tokens of $du_1$ and $du_n$, respectively.

Then, we feed $D$ to the encoder stack (i.e., the encoder of T5) to obtain the encoder's hidden state $Hidden_E$ as follows.

$$Hidden_E = T5 - Encoder(D) \quad (2)$$

Since the state at the $<\backslash s>$ position is not used in T5, we use the Endpoint of the span of the utterance combining the left ($H_i$) and right ($H_{i+1}$) hidden states to represent the corresponding $du_i$, as shown in Eq. 3. Then, we feed the output $V_s^i$ of the span representation layer into the classification layer to judge the topic (T) of the last $du$ as follows, where we choose BiLSTM as our classification layer.

$$V_s^i = Con(H_i, H_{i+1}) \quad (3)$$

$$V_s = [V_s^1, \ldots, V_s^n] \quad (4)$$

$$P = Classifier(V_s) \quad (5)$$

### 3.2 Generation Module

At the decoding stage, we think that solely relying on label data to build the target sentence doesn't fully utilize the generation model's comprehension abilities. There is no additional relevant information in the current corpus. Hence, we pre-process it following previous work. Furthermore, to uncover more information from dialogues, we break it down into two additional levels based on the information already present in the corpus. The dialogue is divided into topic-level and turn-level, considering the topic distribution and speaker information, respectively.

Before initiating the model training, we employ various methods to extract relevant information from dialogues. We describe the details of data pre-processing in the following. Based on the pre-processing results, we formulate the target sentences accordingly. Our model is then segmented into three levels of granularity. Firstly, we translate the original topic labels into the target sentences. Secondly, we analyze the topic information between topic blocks and identified keywords for each topic block which are used to form the target sentences. Lastly, we apply semantic role labeling (SRL) on the dialogue sentences to create the target sentences. In this module, the weights of decoders from different granularities are shared.

**Label-level Prompt** First, we apply the prompt learning to the original topic labels in the dialogue topic corpus. Specifically, we design a target sentence based on the traditional prompting methods and take into account the definition of the task and the semantics of the labels. The target sentence is as follows.

*Chinese: 相对上文，当前话语的话题 [MASK]。*

*English: Relative to the above, the topic of the current discourse has [MASK].*

**Topic-level Prompt** In dialogue, a topic consists of several utterances, which are around the same topic. To enhance the model's performance, we aimed to provide topic information for each topic in addition to building target sentences using label information. However, the corpus does not have such labels. Hence, we utilized keywords extracted for each extracted topic as the topic information for all sentences in the topic's block. Specifically, we merged turns in the same topic block as input, eliminating speaker information. Moreover, the current topic and the previous topic are used to generate topic information, considering the contextual information. Besides, the response utterance is included in the current topic block to utilize the response information.

We created templates to construct target sentences and filled them with topic information, using keyBERT [21] for keyword extraction in each dialogue topic block. For the final set of candidate tuples, we apply these rules for filtering as follows.

- We use the confidence values from keyBERT to choose the top keywords.
- As the best keywords of consecutive blocks may overlap, we prioritize fulfilling the block with the highest-ranked topic of the turn, and then the next block chooses the next best keyword from the candidate tuple.
- For those topic blocks with no candidate tuple, we manually select the keywords for those blocks.

The specific form of the target sentence is as follows, where T2 is the current topic block and T1 is the previous topic block.

*Chinese: 前文在谈论[MASK]，而当前在谈论[MASK]，因此对话话题[MASK]。*

*English: Since the previous text is talking about [MASK] and the current text is talking about [MASK], the topic of conversation has [MASK].*

**Turn-level Prompt** We think the key information for a topic shift is also hinted at various turns besides that obtained at the topic level. To better detect topic shift scenarios, it would be beneficial for the model to acquire semantic information of utterances, such as semantic roles. However, this information is absent in the dialogue topic corpus. Hence, we applied the SRL tool LTP[1] for Chinese and AllenNLP[2] for English to extract all possible tuples for turns (utterances). Then we use the following three rules to filter out those redundant tuples as follows.

- **Streamlining core semantics**. Except for the elements mentioned in Rule 3, we remove the unimportant elements from the extracted candidate tuples.
- **Reducing semantic overlapping**. Since the long nested sentences lead to semantic overlapping, we remove the small tuples that are included in the larger candidate tuples. Finally, considering that the final result is too long to invalidate the target sentence, we limit the length range of the final information to 10 or fewer.
- **Extracting key information**. We tend to select the role A1 (i.e., patient) in SRL as the result in the sentence, followed by the attribute Predicate. And for a few semantically incomplete tuples (i.e., the tuples do not contain A0 and A1 in SRL), we manually determine the final information based on the content of the turn.

The specific form of the target sentence is as follows, where S1 and S2 refer to the created sentence of the previous turn and the current turn, respectively.

---

[1] http://ltp.ai/index.html
[2] https://allennlp.org

*Chinese: 前文在谈论[MASK]，而当前在谈论[MASK]，因此对话话题[MASK]。*
*English: Since the previous text is talking about [MASK] and the current text is talking about [MASK], the topic of conversation has [MASK].*

**Table 1.** Examples of label-level, topic-level, and turn-level target sentences, where T1 and T2 are the key information of the previous topic block Block1 and the current block Block2 respectively, S1 and S2 are the key information of Sentence1 and Sentence2 respectively, and Label represents the topic shift information. For different positions of the [MASK], we comment the corresponding hints in parentheses.

| Attributes | Content |
| --- | --- |
| Template$_{Label}$ | Relative to the above, the topic of the current discourse has [MASK](Label). |
| Block1 | hey ! do you love cats ? hey . . . i am a dog person , i have two. ah that is cool , i have two cats and got a collection of 1000 hats for them ! wow ! ! ! that is a lot lol |
| T1 | cats |
| Block2 | yeah , i have a weakness for cats and vanilla ice cream , they are the best ! my weakness is eating when i am bored |
| T2 | weakness |
| Template$_{Topic}$ | Since the previous text is talking about [MASK](T1) and the current text is talking about [MASK](T2) the topic of conversation has [MASK](Label). |
| Turn1 | yeah , i have a weakness for cats and vanilla ice cream , they are the best ! |
| S1 | a weakness for cats and vanilla ice cream |
| Turn2 | my weakness is eating when i am bored |
| S2 | my weakness |
| Template$_{Turn}$ | Since the previous text is talking about [MASK](S1) and the current text is talking about [MASK](S2), the topic of conversation has [MASK](Label). |

Table 1 shows examples of label-level, topic-level, and turn-level target sentences (i.e., template) in English. These target sentences are constructed to prompt the model to predict the information in the corresponding position of [MASK] and finally determine whether the response (i.e., turn2 in Table 1) is the beginning of a new topic.

Finally, we use the final constructed target sentence T in the same style as the encoding layer as follows.

$$T_l = W_T <\backslash s> (l \in \{Label, Topic, Turn\}) \quad (6)$$

where $W_T = \{w_T^1, ..., w_T^n\}$ denotes the sequence of tokens of the target sentence. Then we feed them to the decoding layer to obtain the decoder hidden state $H_l$ as follows.

$$H_l = T5 - Encoder(T_l) \quad (7)$$

Finally, we use a linear layer generator with softmax to produce the predicted target sentence, where the last [MASK] of the target sentence will be predicted as "SHIFT" or "NON-SHIFT".

### 3.3 Model Training

We learned the above two modules together. The loss function of the classifier ($L_{Class}$) and the multi-granularity generator ($L_{Label}$, $L_{Topic}$, $L_{Turn}$) is the cross-entropy loss, and the total loss (Loss) is the sum of both losses, as follows.

$$Loss = L_{Class} + L_{Label} + L_{Topic} + L_{Turn} \tag{8}$$

## 4 EXPERIMENTATION

### 4.1 Datasets and Experimental Settings

We evaluate our model on two datasets, the Chinese CNTD and English TIAGE. Following previous work, we used the same dataset partitioning on English TIAGE and Chinese CNTD. Based on the dataset of CNTD and TIAGE, we extract (context, response) pairs from each dialogue as input and the label of response as a target for the response-known task. In our experiments, every utterance except the first utterance of the dialogue can be considered as a response. As for evaluation in all experiments of this paper, we report Precision (P), Recall (R), and Macro-F1 scores.

Our experiments are all performed on 3090Ti and use Pytorch and Huggingface as deep learning frameworks, with 2 BiLSTM layers in encoding for both English and Chinese. For Chinese, we use mt5(base) [22] as our T5 model, which is pre-trained on the mC4 corpus, covering 101 languages. It is a T5(base) model with 12 encoder layers and 12 decoder layers. Besides, the model file used by keyBERT [21] is paraphrase-multilingual-MiniLM-L12-v2 for Chinese and paraphrase-MiniLM-L6-v2 for English. For each experiment, we set the batch size to 2 and the number of training epochs to 20. In addition, we used the warm-up strategy as well as the AdamW optimizer and set the decay factor to 0.01.

### 4.2 Experimental Results

In the task of dialogue topic shift detection, Xie et al. [9] is the only work that established a benchmark using the T5 model on TIAGE. Due to the similarity of this task to topic segmentation, we also attempted to utilize the hierarchical model along with the pre-trained model as our baseline for topic shift detection. While in the pre-trained model, T5 is considered the SOTA model. Hence, we conduct the following baselines for comparison: 1) **RoEBRTa** [23], an improvement on BERT; 2) **T5** [9], a modification of the Transformer structure for various NLP tasks as Text-to-Text tasks; 3) **Hier-BERT** [24], a hierarchical structure based on the Transformer model; 4) **BERT+BiLSTM** [24], a combination of BERT for text encoding and a bi-directional LSTM for deep bi-directional language representation; 5) **BERT** [25], a bidirectional encoder based on Transformer for text encoding;

The results are presented in Table 2, which indicate that the pre-trained models show inconsistent performance during the experiments, with RoBERTa exhibiting the poorest results and T5 having the highest performance with a noteworthy F1 score of 81.1. Nevertheless, Compared to a single pre-trained model, it is evident that both hier-BERT and BERT+BiLSTM, which incorporate a hierarchical structure, attain improved performance, recording F1 scores of 81.7 and 82.4, respectively.

The results of the experiments suggest that models incorporating a hierarchical structure provide more consistent results in the task of dialog topic detection. Moreover, our model (Ours) further outperforms the best baseline BERT+BiLSTM significantly ($p < 0.01$), with a 3.0 improvement in F1-score. This result verifies the effectiveness of our proposed model.

**Table 2.** Results of the baselines and ours on CNTD ($p < 0.01$).

| Model | P | R | F1 |
| --- | --- | --- | --- |
| BERT | 82.9 | 79.2 | 80.8 |
| RoBERTa | 84.4 | 75.4 | 78.6 |
| T5 | 83.0 | 79.7 | 81.1 |
| BERT+BiLSTM | 82.8 | 82.0 | 82.4 |
| Hier-BERT | 85.6 | 79.0 | 81.7 |
| Ours | **85.7** | **83.8** | **84.7** |

**Table 3.** Results of the baselines and ours on TIAGE ($p < 0.01$).

| Model | P | R | F1 |
| --- | --- | --- | --- |
| BERT | 68.5 | 65.4 | 66.6 |
| T5 | **76.5** | 72.2 | 73.9 |
| BERT+BiLSTM | 75.8 | 70.8 | 72.7 |
| Hier-BERT | 73.8 | 69.6 | 71.2 |
| Ours | 73.8 | **77.2** | **76.2** |

In addition, we also evaluate our model and the baselines in English TIAGE as shown in Table 3. Compared with BERT, both the hierarchical structure models Hier-BERT and BERT+BiLSTM can obtain better performance. However, different from the results in Chinese, T5 is better than the other three baselines in English. Our proposed model outperforms the best baseline T5 significantly with a 2.3 improvement in F1-score. This result further verifies the effectiveness of our proposed model.

### 4.3 Ablation Study on Classification and Generation

We statistically analyzed the performance of the classification and generation modules in our proposed model. The results are shown in Table 4, where cla and gen indicate the classification module and the generation module, respectively. As shown in Table 4, it is surprising that cla and gen achieve the same precision, recall, and F1 score. This indicates that the generation model can achieve equivalent performance to the classification model. Our proposed model combining classification and generation cla + gen) is better than cla and gen. This result shows the combination of classification and generation is also an effective way for dialogue topic shift detection and these two models can interact and promote each other.

Table 4. Results of the classification and generation on CNTD.

| Model | P | R | F1 |
| --- | --- | --- | --- |
| gen | 83.8 | 81.1 | 82.3 |
| cla | 83.8 | 81.1 | 82.3 |
| gen+cls(Ours) | **85.7** | **83.8** | **84.7** |

### 4.4 Ablation Study on Different Levels of Prompt

The results are shown in Table 5. In the case of the single-level prompt, all the results of label-level, topic-level, and turn-level prompts are better than the basic T5, especially the topic level. This indicates that all three-level prompts are effective for dialogue topic shift detection. Moreover, the performance of the topic level reaches 83.9 in the F1 value and gains the highest improvement (+1.7). It demonstrates that the key information from the topic block has more effective topic information to enhance the model to distinguish different topic shift situations.

Table 5. Ablation experiments at different levels of granularity on CNTD.

| Model | P | R | F1 |
| --- | --- | --- | --- |
| T5 | 84.5 | 80.5 | 82.2 |
| +Label | 82.9 | 81.9 | 82.4 |
| +Topic | 85.4 | 82.6 | 83.9 |
| +Turn | 82.6 | 82.9 | 82.7 |
| +Label+Topic | 84.5 | 82.3 | 83.4 |
| +Label+Turn | 83.3 | 81.7 | 82.5 |
| +Topic+Turn | **86.1** | 83.1 | 84.4 |
| +Label+Topic+Turn | 85.7 | **83.8** | **84.7** |

In addition, it can be noted that both the combination of the label-level and Topic-level prompt (Label + Topic) and the combination of the label-level and Turn-level prompt (Label + Turn) will harm the performance, in comparison with the single-level prompt. This indicates that the information of Label and Topic/Turn is partly crossed and even has a negative impact. It may also be caused by the different forms of target sentences at different granularities. In the case of only two granularities, the different forms of target sentences interact with each other leading to a degradation of performance. In the case of three granularities, the model is dominated by the second form of target sentences, so the performance can be improved. On the contrary, the combination of the Topic-level and Turn-level prompt (Topic + Turn) is better than the single-level prompts Topic and Turn. This indicates that these two prompts can promote each other. Moreover, if we combine all three prompts (Label + Topic + Turn), it can improve the F1 score in comparison with the above combinations.

## 5 Conclusion

In this paper, we introduce a prompt-based model with multi-granularity to detect the topic shift in dialogues, which consists of a classification module and a generation module based on T5. Experimental results on our annotated Chinese dataset CNTD and the publicly available English TIAGE dataset show that the proposed model outperforms the baselines. Further experiments show that the information extracted at different levels of granularity effectively helps the model comprehend the conversation topics. However, when analyzing and observing the information we extracted at different granularities, it is clear that this key information existence of errors and noise. Our future work will focus on improving the reliability of dialogue information mining, and also explore the finer granularity of topic shift scenarios.

## 6 Acknowledgements

The authors would like to thank the three anonymous reviewers for their comments on this paper. This research was supported by the National Natural Science Foundation of China (Nos. 62276177, and 61836007), and Project Funded by the Priority Academic Program Development of Jiangsu Higher Education Institutions (PAPD).